# STUDYING A CHAOTIC SPIKING NEURAL MODEL

Mohammad Alhawarat[1], Waleed Nazih[2] and Mohammad Eldesouki[3]

[1,2]Department of Computer Science, College of Computer Engineering and Sciences,
Salman Bin Abdulaziz University, Kingdom of Saudi Arabia
[3]Department of Information Systems, College of Computer Engineering and Sciences,
Salman Bin Abdulaziz University, Kingdom of Saudi Arabia

*ABSTRACT*

*Dynamics of a chaotic spiking neuron model are being studied mathematically and experimentally. The Nonlinear Dynamic State neuron (NDS) is analysed to further understand the model and improve it. Chaos has many interesting properties such as sensitivity to initial conditions, space filling, control and synchronization. As suggested by biologists, these properties may be exploited and play vital role in carrying out computational tasks in human brain. The NDS model has some limitations; in thus paper the model is investigated to overcome some of these limitations in order to enhance the model. Therefore, the model's parameters are tuned and the resulted dynamics are studied. Also, the discretization method of the model is considered. Moreover, a mathematical analysis is carried out to reveal the underlying dynamics of the model after tuning of its parameters. The results of the aforementioned methods revealed some facts regarding the NDS attractor and suggest the stabilization of a large number of unstable periodic orbits (UPOs) which might correspond to memories in phase space.*

## 1. INTRODUCTION

Biologists suggest that chaos may play an important role in human brain [1-8]. Chaos has many interesting properties that might be exploited in carrying out information processing tasks. Such properties are: sensitivity to initial conditions, space filling, control, synchronization and rich dynamics that can be accessed using different control methods. When Artificial Neural Networks (ANNs) are equipped with chaos they might provide an access to large number of rich dynamic behaviours. These can be accessed if appropriate control mechanisms is chosen such as feedback control [9-11]. If such approaches are applied to chaotic neural models then the model might stabilize into one of many UPOs that a chaotic attractor encompasses.

In recent years, many chaotic neural models have been devised to study how such rich dynamic behaviours might be exploited in carrying out information processing tasks. One of these models is the NDS model [12] which is based on Rössler system [13]. It is a simple chaotic system with one nonlinear term. Rössler system has been studied in different contexts such as control and biology to name a few [14,15].

The NDS model was first devised in 2003 [16] and it represents a chaotic spiking neural model. The authors introduced the model in [16] to represent an infinite state machine using the rich dynamic behaviours that model encompasses. The authors have demonstrated that using input signals along with a control mechanism could stabilize the model attractor to one of its UPOs. The control mechanism used was a modified version of Pyragas [11]. The authors have studied small networks of 2-3 neurons, the results shows that the network has been stabilized to an UPO according to a periodic length that is relative to the input pattern.

Using only one NDS neuron, its attractor might have access to large number of UPOs that can be stabilised to. These might be mapped to memories in phase space. In contrast, a Hopfield neural network can give only 0.15n memory size (where n is the number of neurons in the network).
The NDS model has been studied in [12, 17-20]. In [17] Lorenz attractor has been used instead of Rössler in an application of human motion detection from a sequence of video frames. In another work [19] the authors suggested that chaos may equip mammalian brain with a mechanism that helps in solving hard nonlinear problems. In [20] networks of NDS neurons have been studied in terms of Spike Time Dependent Plasticity (STDP) which is a property of cortical neurons. Experiments results suggested that NDS neurons may have the propriety of the realism of biological neural networks. In [18] the NDS model has been investigated thoroughly. The author has studied the chaotic behaviour of the model both experimentally and analytically. The study has shown interesting results.

This paper is organized as follows: in section 2 the original Rössler model is described, in section 3 the NDS model is introduced, in section 4 the discretization method that has been used to devise the NDS model is analysed and studied, in section 5 the experimentation setups that are carried out to tune the parameters of the NDS model are described, section 6 is devoted to the analysis of the dynamics of the NDS attractor, section 7 includes discussions and finally section 8 concludes the paper.

## 2. RÖSSLER CHAOTIC ATTRACTOR

Rössler system [13] is a simple chaotic dynamical system and is represented by the following equations:

$$x' = -y - u \qquad (1)$$
$$y' = -x - a*y \qquad (2)$$
$$z' = b + z(x - c) \qquad (3)$$

Where *a* and *b* are usually fixed and c is called the control parameter. The usual parameter settings for the Rössler attractor are $a = 0.2$, $b = 0.2$, and $c = 5.7$, and the attractor for such settings is shown in figure 1.

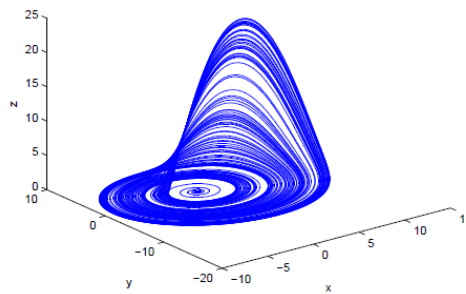

Fig. 1 The Rössler chaotic attractor with parameters $a = 0.2$, $b = 0.2$, and $c = 5.7$.

## 3. DESCRIBING THE NDS MODEL

The NDS model is first proposed in [12]. The NDS neuron is a discretized model that is based on Rössler's attractor [13] as described by equations 1-3 in section 2.

Large number of orbits with varying periodicity might be stabilised by varying different system's parameters such as time delays, period length $\tau$ and initial conditions.

The NDS model simulates a novel chaotic spiking neuron and is represented by:

$$x(t+1) = x(t) + b(-y(t) - u(t)) \quad (4)$$

$$y(t+1) = y(t) + c(x(t) + ay(t)) \quad (5)$$

$$u(t+1) = \begin{cases} \eta_0 & u(t) > \theta \\ u(t) + d(v + u(t)(-x(t)) + ku(t)) + F(t) + In(t) & u(t) \leq \theta \end{cases} \quad (6)$$

$$F(t) = \sum_{j=1}^{n} w_j \gamma(t - \tau_j) \quad (7)$$

$$In(t) = \sum_{j=1}^{n} I_j(t) \quad (8)$$

$$\gamma(t+1) = \begin{cases} 1 & u(t) > \theta \\ 0 & u(t) \leq \theta \end{cases} \quad (9)$$

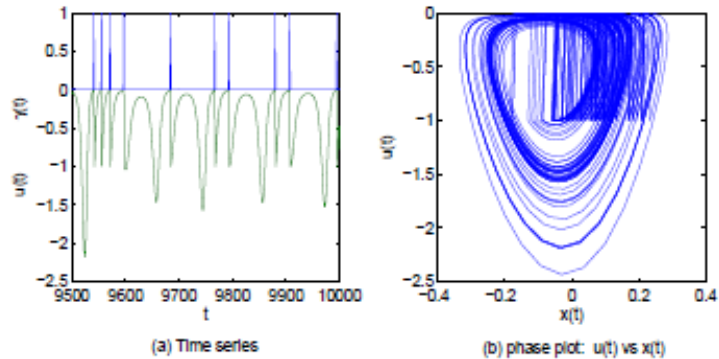

(a) Time series  (b) phase plot: u(t) vs x(t)

Fig. 3 The chaotic behaviour of a NDS neuron without input (a) the time series of $u(t)$ and $\gamma(t)$ and (b) the phase space of $x(t)$ versus $u(t)$

where $x(t)$, $y(t)$ and $u(t)$ describe the internal dynamics of the model, $\gamma(t)$ is the model's binary output, $F(t)$ is the feedback signal, $In(t)$ represents the external input spike train, and the parameters of the model are: $a = 0.002$, $v = 0.002$, $b = 0.03$, $c = 0.03$, $d = 0.8$, $k = -0.057$, $\theta = -0.01$, $\eta_0 = -0.7$ and $\tau_j$ is the period length of the feedback signal.

The discretization method that is used in constructing the NDS model has been carried out by scaling the system variables $x(t)$, $y(t)$ and $u(t)$ using different scaling constants: $b$, $c$, $d$ which been tuned by carrying out experiments until major dynamics of the Rössler system are preserved.

Figure 3 depicts the dynamics of a single NDS neuron without input, whereas figure 4 shows the NDS dynamics when it is stabilised to period-4 orbit due to the feedback control mechanism $F$ and a time delayed feedback connection.

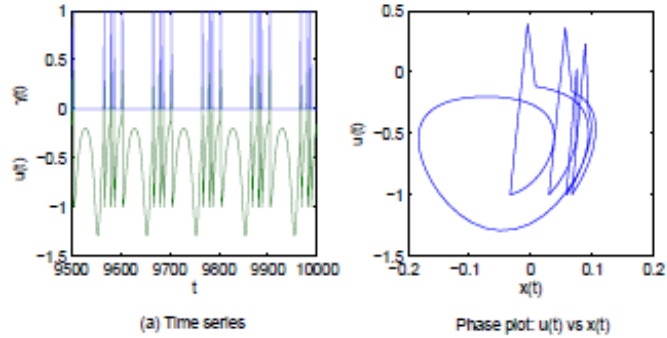

Fig. 4 The stabilizing of period-4 orbit of a NDS neuron with feedback connection.

## 4. DISCRETIZATION METHOD

There are different discretization methods that are used to convert a continuous system into discrete. One of the nonstandard methods is Euler's Forward differentiation that is used in formulating simple simulation systems. A time step $TS$ is used in calculating the approximate value of next step of a continuous system. For example, when Euler's Forward differentiation method is applied to equation 1 then it will become:

$$\dot{x}(t_k) = \frac{x(t_{k+1}) - x(t_k)}{TS} = -y(t_k) - u(t_k) \qquad (10)$$

and then solving for $x(t_{k+1})$:

$$x(t_{k+1}) = x(t_k) + TS(-y(t_k) - u(t_k)) \qquad (11)$$

if $x(t)$ is used instead of $x(t_k)$, then the equation becomes:

$$x(t+1) = x(t) + TS(-y(t) - u(t)) \qquad (12)$$

Note that $TS$ is chosen to be parameter $b$ with the value 0.03 when equation 12 is compared to equation 4.

For simulation purposes $TS$ is usually chosen to be small and it is preferable to be chosen according to:

$$TS \leq \frac{0.1}{|\lambda|_{max}} \qquad (13)$$

Where $|\lambda|_{max}$ is the largest absolute eigenvalues for the system under study. For the NDS model and according to the mathematical analysis results that is achieved in [18], $|\lambda|_{max} = 5.68698$.

When substituting this value in equation 13:

$$TS \leq 0.0176 \qquad (14)$$

If this value is compared with $b$ it is obvious that the time step that has been chosen doesn't follow the simulation preferable setup.

Also, it is clear from equation 5 that *TS* is chosen to be $c = 0.03$ and a scaling factor is applied to *y* variable using $a = 0.002$.

For the u variable *TS* is chosen to be d =0.8. Moreover, authors in [12] have scaled down constant *c* to 0.057 and modified its sign. Finally, sign of *x* and *k* have also been changed.

The aforementioned modifications have affected the original Rössler system properties in phase space and in Eigen space as well. According to [18], the two spiral saddle points of fixed points of original Rössler system have become two spiral repellors in the NDS model. This major change might be due to the change to both scaling factors and sign of *x* and *k*.

The change in the fixed points indicate that the NDS model have weak connection to the original Rössler system. This conclusion was shown in [18] when the author indicated that UPOs exist only as the results of both the acting forces of the spiral repellors of the attractor along with the applied reset mechanism. Without the reset mechanism the spiral repellors will enforce any nearby trajectory to approach infinity.

When Euler's forward differentiation method is applied to Rössler system then equations 1- 3 become:

$$x(t+1) = x(t) + TS(-y(t) - u(t)) \qquad (15)$$
$$y(t+1) = y(t) + TS(x(t) + ay(t)) \qquad (16)$$
$$u(t+1) = u(t) + TS(b + z(t)(x(t) - c)) \qquad (17)$$

Where $TS = 0.0055$, $a = 0.2$, $b = 0.2$, and $c = 5.7$.

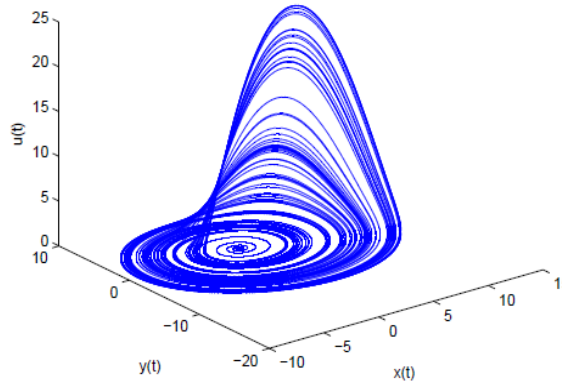

Fig. 5 The Discrete version of Rössler attractor based on equations 15- 17

An experiment setup has been prepared to make sure the new equations will preserve the original Rössler system attractor.

The following changes need to be made to equations 15-17 to follow the NDS equations:
1. Set TS to 0.0055.
2. Change *b and c* in equation 17 to *v and k*.
4. Change *TS* in equations 15,16 and 17 to *b,c* and *d* and set them to the value of *TS*.
7. Change the value of *a* and *v* and to 0.2.
8. Change the value of *k* to 5.7
9. Rename variable *z* to be *u*.
10. Change the sign of the term *(x(t)−k)* in equation 17 to become *(−x(t)+k)*

When applying the previous changes, then equations 15- 17 become:

$$x(t+1) = x(t) + b(-y(t) - u(t)) \qquad (18)$$
$$y(t+1) = y(t) + c(x(t) + ay(t)) \qquad (19)$$
$$u(t+1) = u(t) + d(v + u(t)(-x(t) + k)) \qquad (20)$$

Where $a = 0.2$, $b = c = d = 0.0055$, and $k = 5.7$.

To verify the new model an experiment has been carried out to depict the attractor that represents equation 18-20. This is depicted in figure 6. Note that the original Rössler attractor has disappeared.

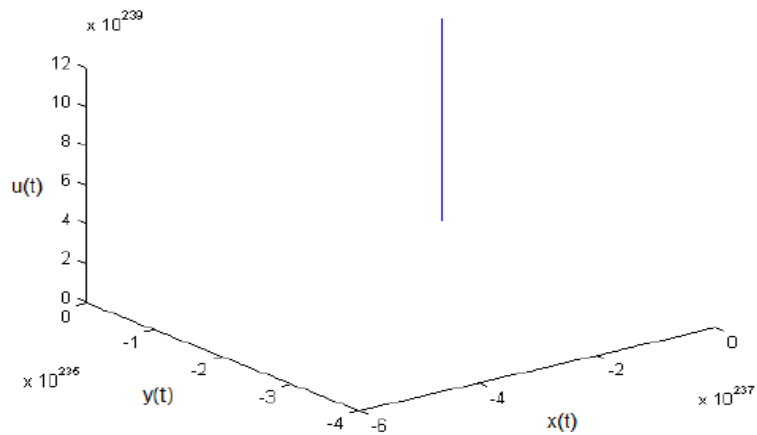

Fig. 6 The attractor of the model that is based on equations 18- 20.

## 5. TUNING THE PARAMETERS OF THE NDS MODEL

In this section the details of tuning the parameters of the NDS model is shown. To achieve this, many experiment setups have been prepared. The parameters that considered are $a,v,b,c,d$ and $k$, the other parameters, viz., $\theta, \eta_0, \tau_j$ have already been studied in [18]. In all experiments one NDS neuron is used where the feedback control is applied after time step of 1000. After that the experiment runs for 9000 iterations.

Wide range of settings have been chosen. Values of the variables of the model are recorded and then are depicted in phase space. Values of the parameters are considered valid if an attractor exists for each setting. The results of valid ranges for the model parameters are shown in table 1.

Table 1 Parameter value's ranges

| Parameter | a,v | b,c | D | K |
|---|---|---|---|---|
| Value | 0.001-0.1 | 0.01-0.055 | 0.8-0.9 | -(0.055-0.058) |

Different parameter settings have been chosen as stated in table 2. The NDS original parameter setup is used in setup 7 to compare this setup with other settings to highlight areas of enhancements of the model.

Table 2 Parameter settings with different selected values from the ranges appear in table1

| Parameter | a,v | b,c | D | K |
|---|---|---|---|---|
| **Setup 01** | 0.001 | 0.03 | 0.8 | -0.057 |
| **Setup 02** | 0.01 | 0.03 | 0.8 | -0.057 |
| **Setup 03** | 0.1 | 0.03 | 0.8 | -0.057 |
| **Setup 04** | 0.002 | 0.001 | 0.8 | -0.057 |
| **Setup 05** | 0.002 | 0.02 | 0.8 | -0.057 |
| **Setup 06** | 0.002 | 0.05 | 0.8 | -0.057 |
| **Setup 07** | 0.002 | 0.03 | 0.8 | -0.057 |
| **Setup 08** | 0.002 | 0.03 | 0.85 | -0.057 |
| **Setup 09** | 0.002 | 0.03 | 0.9 | -0.057 |
| **Setup 10** | 0.002 | 0.03 | 0.8 | -0.055 |
| **Setup 11** | 0.002 | 0.03 | 0.8 | -0.056 |
| **Setup 12** | 0.002 | 0.03 | 0.8 | -0.058 |
| **Setup 13** | 0.01 | 0.05 | 0.85 | -0.055 |
| **Setup 14** | 0.002 | 0.015 | 0.8 | -0.058 |
| **Setup 15** | 0.1 | 0.04 | 0.8 | -0.056 |

Another experiment setup has been prepared to verify the capacity of the attractor in terms of UPOs that it might encompasses. The average number of successfully stabilized UPOs is computed over 1000 run according to the parameter settings that appear in table 2 and then depicted in figure 7.

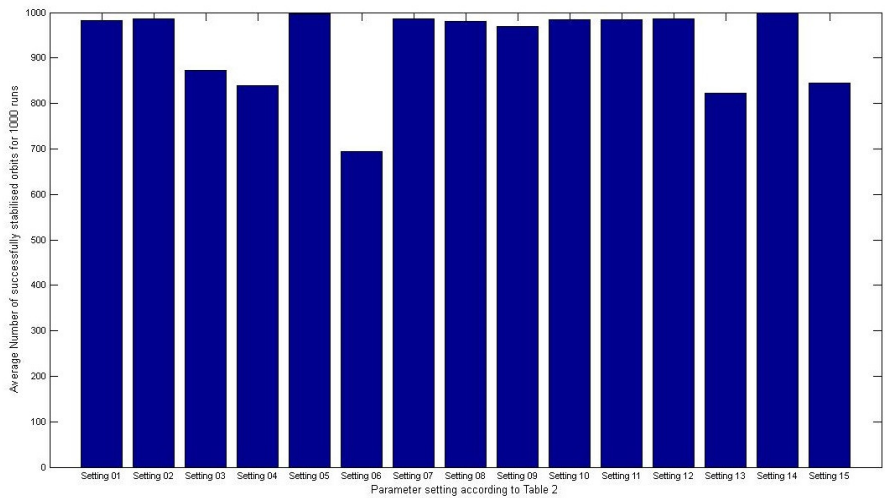

Fig. 7 Average number of Stabilized UPOs over 1000 run based on the parameter settings that appear on table 2.

Figure 7 suggests that slightly better parameter settings exist such as Settings 14 and Settings 05 increased the attractor capacity when compared to the NDS original settings (Setting 07).

## 6. EIGEN SPACE ANALYSIS

In this section the fixed points of the NDS attractor will be calculated for the different parameter settings that appear in table 2. Also Eigen space will be computed and compared to the NDS and original Rössler attractors. Before that a gentle introduction to Eigen space terminology will be prefaced. For more information about nonlinear dynamics and mathematical details used in this section please refer to [21-28].

### 6.1 Fixed points and Eigen space Analysis

In a chaotic system there are poles that organize and control the behaviour of the trajectories that pass nearby them, and are known as fixed points. Around each of these points there is a basin of attraction or repulsion. Once a trajectory enters one of these basins of attraction its behaviour will be affected, then it will either be attracted and end up in a periodic or fixed point behaviour or it will be repelled and move to another basin of attraction. These fixed points restrict the behaviour of trajectories of a chaotic system in a fractal-bounded area and determine the shape of the attractor in the phase space.

The types of fixed points of a dynamic system can be determined by finding its characteristic values. These characteristic values are then used to analyze and understand the different system behaviours around the fixed points including the stability of these points. The types of fixed points available depend on the dimensions of the dynamic system under study.

Finding and analyzing fixed points in three-dimensional state space is more complicated than in one and two-dimensional state space. That is because systems in three-dimensional state space require:
- Solving cubic equations which is usually hard and sometimes is impractical.
- A larger Jacobian matrix which requires more calculations.
- Calculating the three characteristic values that are used to determine the nature of each fixed point of the system.

Finding the characteristic values from the Jacobian matrix includes solving quadratic equations. There are three roots that result from solving cubic equations, which are the characteristic values. The combination of these characteristic values together determines the type of the system fixed point(s).

Then the characteristic values could be in one of these categories:
- *All real numbers:*
  - All negative. The fixed point is of type Node: All trajectories head directly to it.
  - All positive. The fixed point is of type Repellor: All trajectories move away directly from it.
  - Two negative and one positive. The fixed point is of type Saddle point Index-1: Trajectories attracted by the saddle point on a surface (two stable manifolds) and repelled on a curve (one unstable manifold).
  - Two positive and one negative. The fixed point is of type Saddle point Index-2: Trajectories repelled by the saddle point on a surface (two unstable manifolds) and attracted on a curve (one stable manifold).

The index number here indicates the number of the unstable manifolds.
- *Mixed real and imaginary numbers:*
  - All negative; one real and two complex conjugate pair. The fixed point is of type Spiral Node.

- o All positive; one real and two complex conjugate pair. The fixed point is of type Spiral Repellor.
- o Two negative (complex conjugate pair) and one real positive. The fixed point is of type Spiral Saddle point Index-1: Trajectories spiral around the saddle point while they are attracted by the two stable manifolds.
- o Two positive (complex conjugate pair) and one real negative. The fixed point is of type Spiral Saddle point Index-2: Trajectories spiral around the saddle point on a surface (two unstable manifolds) while they are moving away from the saddle point.

### 6.2 Calculating Fixed points and Eigen space values

According to [18], the two fixed points for the NDS model are of type spiral repellor, where the original Rössler system fixed points are of type spiral saddle point.

An extensive mathematical calculations need to be carried out to reveal the effect of parameter tuning that has been discussed in the previous section. Then an experiment setup has been prepared to calculate the fixed points and the Eigen space for the different parameter settings according to table 2.

First the fixed points have been calculated for each parameter setup by solving the quadratic equations of the system. Second, the Jacobian matrices have been belt for each equilibrium vector, and then the Eigen vectors and Eigen values have been calculated. Finally, the fixed point type is specified based on the classification that is aforementioned in subsection 6.1. The results of running the experiment show different fixed points and Eigen values as shown in tables 4, 5 and 6.

Table 4 Fixed points for each parameter settings that appears in table 2.

| Fixed Points | *First Fixed Point* | | | *Second Fixed Point* | | |
|---|---|---|---|---|---|---|
| | X | y | Z | x | y | z |
| **Setup 01** | -0.05702 | 57.01754 | -57.01754 | 0.00002 | -0.01754 | 0.01754 |
| **Setup 02** | -0.05870 | 5.87035 | -5.87035 | 0.00170 | -0.17035 | 0.17035 |
| **Setup 03** | -0.13248 | 1.32482 | -1.32482 | 0.07548 | -0.75482 | 0.75482 |
| **Setup 04** | -0.05707 | 28.53504 | -28.53504 | 0.00007 | -0.03504 | 0.03504 |
| **Setup 05** | -0.05707 | 28.53504 | -28.53504 | 0.00007 | -0.03504 | 0.03504 |
| **Setup 06** | -0.05707 | 28.53504 | -28.53504 | 0.00007 | -0.03504 | 0.03504 |
| **Setup 07** | -0.05707 | 28.53504 | -28.53504 | 0.00007 | -0.03504 | 0.03504 |
| **Setup 08** | -0.05707 | 28.53504 | -28.53504 | 0.00007 | -0.03504 | 0.03504 |
| **Setup 09** | -0.05707 | 28.53504 | -28.53504 | 0.00007 | -0.03504 | 0.03504 |
| **Setup 10** | -0.05507 | 27.53632 | -27.53632 | 0.00007 | -0.03632 | 0.03632 |
| **Setup 11** | -0.05607 | 28.03567 | -28.03567 | 0.00007 | -0.03567 | 0.03567 |
| **Setup 12** | -0.05807 | 29.03444 | -29.03444 | 0.00007 | -0.03444 | 0.03444 |
| **Setup 13** | -0.05676 | 5.67617 | -5.67617 | 0.00176 | -0.17617 | 0.17617 |
| **Setup 14** | -0.05807 | 29.03444 | -29.03444 | 0.00007 | -0.03444 | 0.03444 |
| **Setup 15** | -0.13185 | 1.31846 | -1.31846 | 0.07585 | -0.75846 | 0.75846 |

Table 5 Eigen values for the *first fixed point* and the corresponding fixed point type for each parameter setting that appears in table 2.

|  | Eigen Values *For the First Fixed Point* |  |  | *Fixed Point Type* |
| --- | --- | --- | --- | --- |
| **Setup 01** | 1.0000+1.1702i | 1.0000-1.1702i | 1.0000 | Spiral Repellor |
| **Setup 02** | 1.0007+0.3765i | 1.0007-0.3765i | 1.0003 | Spiral Repellor |
| **Setup 03** | 1.0294+0.1782i | 1.0294-0.1782i | 1.0046 | Spiral Repellor |
| **Setup 04** | 1.0000+0.1511i | 1.0000-0.1511i | 1.0000 | Spiral Repellor |
| **Setup 05** | 1.0000+0.6760i | 1.0000-0.6760i | 1.0000 | Spiral Repellor |
| **Setup 06** | 1.0000+1.0695i | 1.0000-1.0695i | 1.0001 | Spiral Repellor |
| **Setup 07** | 1.0000+0.8281i | 1.0000-0.8281i | 1.0001 | Spiral Repellor |
| **Setup 08** | 1.0000+0.8535i | 1.0000-0.8535i | 1.0001 | Spiral Repellor |
| **Setup 09** | 1.0000+0.8783i | 1.0000-0.8783i | 1.0001 | Spiral Repellor |
| **Setup 10** | 1.0000+0.8135i | 1.0000-0.8135i | 1.0001 | Spiral Repellor |
| **Setup 11** | 1.0000+0.8208i | 1.0000-0.8208i | 1.0001 | Spiral Repellor |
| **Setup 12** | 1.0000+0.8353i | 1.0000-0.8353i | 1.0001 | Spiral Repellor |
| **Setup 13** | 1.0007+0.4937i | 1.0007-0.4937i | 1.0005 | Spiral Repellor |
| **Setup 14** | 1.0000+0.5905i | 1.0000-0.5905i | 1.0000 | Spiral Repellor |
| **Setup 15** | 1.0293+0.2069i | 1.0293-0.2069i | 1.0061 | Spiral Repellor |

Table 6 Eigen values for the *second fixed point* and the corresponding fixed point type for each parameter setting that appears in table 2.

|  | Eigen Values *For the Second Fixed Point* |  |  | *Fixed Point Type* |
| --- | --- | --- | --- | --- |
| **Setup 01** | 1.0031+0.0280i | 1.0031-0.0280i | 0.9483 | Spiral Repellor |
| **Setup 02** | 1.0209+0.0075i | 1.0209-0.0075i | 0.9116 | Spiral Repellor |
| **Setup 03** | 1.0842 | 1.0091 | 0.8038 | Repellor |
| **Setup 04** | 1.0003+0.0009i | 1.0003-0.0009i | 0.9537 | Spiral Repellor |
| **Setup 05** | 1.0045+0.0177i | 1.0045-0.0177i | 0.9453 | Spiral Repellor |
| **Setup 06** | 1.0070+0.0433i | 1.0070-0.0433i | 0.9405 | Spiral Repellor |
| **Setup 07** | 1.0058+0.0262i | 1.0058-0.0262i | 0.9428 | Spiral Repellor |
| **Setup 08** | 1.0060+0.0262i | 1.0060-0.0262i | 0.9396 | Spiral Repellor |
| **Setup 09** | 1.0061+0.0263i | 1.0061-0.0263i | 0.9365 | Spiral Repellor |
| **Setup 10** | 1.0061+0.0259i | 1.0061-0.0259i | 0.9439 | Spiral Repellor |
| **Setup 11** | 1.0059+0.0260i | 1.0059-0.0260i | 0.9433 | Spiral Repellor |
| **Setup 12** | 1.0057+0.0263i | 1.0057-0.0263i | 0.9423 | Spiral Repellor |
| **Setup 13** | 1.0292+0.0180i | 1.0292-0.0180i | 0.8939 | Spiral Repellor |
| **Setup 14** | 1.0036+0.0135i | 1.0036-0.0135i | 0.9464 | Spiral Repellor |
| **Setup 15** | 1.1010 | 1.0123 | 0.7852 | Repellor |

### 6.3 Analyzing the results

It is obvious from table 4 that setup 1, 2, 3, 13 and 15 are affected by tuning the parameters. Some values of the fixed points for the other setups have been slightly changed, but these have not affected the types of the fixed points as shown in both tables 5 and 6.

The change in the values of the fixed points for setup 1, 2, 3, 13 and 15 have affected the type of the fixed points for setup 3 and 15 only as shown in table 6. However, the change from spiral repellor to repellor will not affect the dynamic behaviour of the attractor as both of them will repel trajectories heading toward them in all directions: *x*, *y* and *z*.

## 7. DISCUSSION

There results attained in this paper suggest that there are weak connections between the NDS model and original Rössler system. The discretization method that is used in constructing the NDS model doesn't follow known methods that are used in simulating continuous systems. The authors in [12] has used different time steps and nested scaling factors. These and the change that is made to the sign of both x and k has led to a change in attractor properties such as fixed points. Tuning the parameters of the NDS model has led to slight changes in the capacity of the attractor in terms of the number of stabilized UPOs. The mathematical analysis of the dynamics of the model for different parameter's settings has shown that the fixed point's types remained unchanged.

The two spiral repellors of the NDS model force any nearby trajectory to evolve to infinity, but when NDS model is equipped with a suitable reset mechanism then trajectories will remain in a fractal dimension as shown in figure 3. This is one of the main chaos properties. It has already proven that the NDS model is chaotic according to the Lyapunov exponent estimates obtained in [29]. In contrast, the two spiral saddle points of the Rössler attractor will keep nearby trajectories evolving between the forces of attracting sides of both of saddle points.

When comparing the attractors of both NDS and original Rössler as depicted in figure 1 and figure 3, it is clear that the shape of the attractor is preserved in *x-u* plot except that the attractor in the NDS model is flipped due to the sign change in *x* and *k*. However, when comparing the fixed points, eigenvalues and eigenvectors it is obvious that they are different.

The results obtained in this paper also suggest that both the reset mechanism and the feedback signal are vital for the NDS attractor to exist.

## 8. CONCLUSION

Chaotic spiking neural models have been studied and investigated in recent years to explore the possibilities of exploiting dynamics of such models in carrying out information processing tasks. The NDS model encompasses a large number of UPOs that are stabilised using a feedback control mechanism. Although the NDS model has weak connections to the original Rössler system, still much rich dynamic behaviours are inherited from the Rössler system which can be noticed when comparing figure 1 and figure 3.

The discretization method that is used in devising the NDS model along with the change in sign in the term $u(t)(x(t)-k)$ has affected the shape and properties of the NDS attractor when compared with Rössler system.

Different experiments have been carried out to tune parameters of the NDS model. The valid ranges of the model have not changed the properties of the NDS model both in phase space and Eigen space. The mathematical analysis of the attractor for different parameter settings revealed that the underlying dynamics of the attractor remained the same.

Although there are weak connections between the NDS and the Rössler models, the NDS attractor encompasses large number of UPOs and wide range of dynamic behaviours that may be exploited to carry out information processing tasks.

## ACKNOWLEDGEMENTS

I would like to thank the deanship of scientific research at Salman Bin Abdulaziz University in KSA for supporting this work under grant number 40/H/1432.

**Authors**

Mohammad Alhawarat has got his Ph.D. in chaotic neural networks from School of Technology of Oxford Brookes University, United Kingdom, in 2007. He has worked as an assistant professor in Petra University in Jordan for 1 year, he then joined the College of Computer Engineering and Sciences of Salman Bin Abdulaziz University since 2008 as an assistant professor of computer science. His research interests are chaotic neural networks, data mining and machine learning including natural language processing. 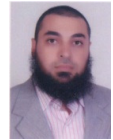

Waleed Nazih has got his M.Sc. degree of Information Technology from Faculty of Computers & Information, Cairo University, 2007. He has worked as lecturer in the College of Computer Engineering and Sciences of Salman Bin Abdulaziz University since 2008. His research interests includes machine learning including speech recognition and natural language processing.

Mohammad Eldesouki has got his M.Sc. in Information Systems from the Faculty of Computers and Information, Helwan University, 2002. He has worked as lecturer at the department of information systems, College of Computer Engineering and Sciences of Salman Bin Abdulaziz University since 2008. His research interests are databases and data mining. 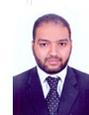